\documentclass{bmvc2k}
\usepackage{multirow}
\usepackage{pgfplots}
\usepackage{wrapfig}
\usepackage{enumitem} 

\title{An Efficient 3D CNN for Action/Object Segmentation in Video}

\addauthor{Rui Hou}{houray@gmail.com}{1}
\addauthor{Chen Chen}{chen.chen@uncc.edu}{2}
\addauthor{Rahul Sukthankar}{sukthankar@google.com}{3}
\addauthor{Mubarak Shah}{shah@crcv.ucf.edu}{4}

\addinstitution{Niantic Labs \\ Sunnyvale, CA, USA}
\addinstitution{University of North Carolina at Charlotte \\ Charlotte, NC, USA}
\addinstitution{Google Research \\ Mountain View, CA, USA}
\addinstitution{University of Central Florida \\ Orlando, FL, USA}

\runninghead{Hou \etal}{An Efficient 3D CNN for Action/Object Segmentation in Video}

\def\eg{\emph{e.g}\bmvaOneDot}

\def\etal{\emph{et al}\bmvaOneDot}
\def\ie{\emph{i.e}\bmvaOneDot}
\begin{document}

\maketitle

\begin{abstract}
 
Convolutional Neural Network (CNN) based image segmentation has made great progress in recent years. However, video object segmentation remains a challenging task due to its high computational complexity. Most of the previous methods employ a two-stream CNN framework to handle spatial and motion features separately. In this paper, we propose an end-to-end encoder-decoder style 3D CNN to aggregate spatial and temporal information simultaneously for video object segmentation. 
%The proposed approach leverages 3D separable convolutions and drastically reduces the number of trainable parameters. 
%Instead of the popular two-stream framework, we adopt 3D CNN to aggregate spatial and temporal information. 
To efficiently process video, we propose 3D separable convolution for the pyramid pooling module and decoder, which dramatically reduces the number of operations while maintaining the performance. Moreover, we also extend our framework to video action segmentation by adding an extra classifier to predict the action label for actors in videos.
Extensive experiments on several video datasets demonstrate the superior performance of the proposed approach for action and object segmentation compared to the state-of-the-art. %\MS {Do you want to mention by how much you improve the state of art?}.

%\MS{Is separability the only novelty? Do you want to mention 3D pyramid, dilated convolution etc. Abstract is too short, unimpressive.}
%\MS {How about action segmentation? That's is mentioned in the title. Has anyone done the way you are doing action segmentation: foreground/background first then classification, may be mention that} . 
\end{abstract}

%-------------------------------------------------------------------------
\section{Introduction}
\label{sec:introduction}
Video object segmentation is a fundamental task in video content analysis. It aims to assign a foreground/background label for each pixel in a video frame. Compared to image segmentation, there are two major differences in video object segmentation.
First, the amount of data to be processed in video can be orders of magnitude greater than in image segmentation, placing greater constraints in terms of computational resources.
Second, the temporal domain provides additional information about object  motion that can judiciously be exploited to improve segmentation performance.

Video object segmentation approaches can be divided into two categories -- semi-supervised and unsupervised. The semi-supervised approaches \cite{Maninis2018Video,Voigtlaender2017Online} assume the foreground object in the first frame of test video is provided and the task is to segment the specified object in the following frames. On the other hand, the unsupervised approaches \cite{khoreva2016learning,caelles2017one,Song_2018_ECCV,tokmakov2017learning} segment foreground objects without any prior knowledge, which is more suited for practical use.  

In this paper, we approach the video object segmentation in the \textbf{unsupervised setting}. We propose an end-to-end encoder-decoder style 3D CNN based method to solve the video object segmentation problem efficiently. Its encoder is composed of a R2plus1D (R2P1D) network \cite{r2plus1d_cvpr18} and a 3D pyramid pooling module. Its decoder is designed to recover both spatial and temporal dimensions to generate an output of the same size as the input clip. Instead of the popular two-stream framework, we adopt 3D CNN to aggregate spatial and temporal information. To efficiently process video, we propose 3D separable convolution for the pyramid pooling module and decoder, which dramatically reduces the number of operations while maintaining the performance. Additionally, we also extend our framework to video action segmentation by adding an extra classifier to predict the action label for actions in videos.
The main contributions of this paper are three-fold:
\begin{enumerate}%[topsep=0pt,itemsep=2pt,parsep=2pt]
\item We propose a simple yet efficient 3D CNN framework for action/object segmentation in videos. To the best of our knowledge, this is the first time 3D CNN has been explored for video object segmentation to simultaneously model the spatial and temporal information in a video.
\item We evaluate our method on video object segmentation and action segmentation benchmarks and demonstrate state-of-the-art performance.
\item We conduct a detailed ablation study to identify the relative contributions of the individual components.
\end{enumerate}

\section{Related Work}
\label{sec:related_work}
Convolutional neural networks have been demonstrated to achieve excellent results in video action understanding \cite{lecun2015deep, karpathy2014large, lstm_ng}. Video should not be treated as a set of independent frames, since the connection between frames provides extra temporal information for understanding. Simonyan \etal \cite{2stream_cnn_simonyan_2014two} propose the two-stream CNN approach for action recognition, which consists of two CNNs taking image and optical flow as input respectively. To avoid computing optical flow separately, Tran \etal \cite{c3d} propose 3D CNN for large scale action recognition. Hara \etal \cite{hara2018can} apply 3D convolution on ResNet structure. Carreira \etal \cite{carreira2017quo} propose I3D by extending Inception network from 2D to 3D and including an extra optical flow stream. Tran \etal \cite{r2plus1d_cvpr18} and Xie \etal \cite{xie2018rethinking} factorize 3D CNN to treat spatial and temporal information separately to reduce the computational cost while keeping the performance. However, to the best of our knowledge, we are the first ones to exploit 3D CNN for video object segmentation.

\textbf{CNN-based segmentation.} The success of CNN-based approaches for image classification \cite{krizhevsky2012imagenet, he2016deep} have led to dramatic advances in image segmentation \cite{long2015fully, noh2015learning}. Many of the segmentation approaches leverage recognition models trained on ImageNet and replace the fully-connected layers with 1$\times$1 kernel convolutions to generate dense (pixel-wise) labels. Recently, the encoder-decoder style network architecture, such as SegNet \cite{badrinarayanan2015segnet} and U-Net \cite{ronneberger2015u}, has been the main stream design for semantic segmentation. Moreover, pyramid pooling \cite{zhao2017pyramid, chen2017rethinking} and dilated convolution \cite{yu2015multi, chen2017rethinking, chen2018deeplab} are effective techniques to improve the segmentation accuracy. Our 3D CNN also builds upon the encoder-decoder structure for video object segmentation. 

\textbf{Video object segmentation.} Video object segmentation \cite{ke2007event, yan2008learning} aims to delineate the foreground object(s) from the background in each frame. Semi-supervised segmentation pipelines \cite{Maninis2018Video,Voigtlaender2017Online} assume that the segmentation mask of the first frame in the sequence during testing is given, 
%and the object maps in video are obtained by %fine-tuning 
%a CNN model. % using the initial mask as 
%\MS {I am not sure this is true, CNN model is not fine tuned, it remains fixed, but segmentation in the first frame is used as extra source of information}. 
%Most of the semi-supervised video object segmentation algorithms 
and exploit temporal consistency in video sequences to propagate the initial segmentation mask to subsequent frames.

In the more challenging unsupervised setting, as we address in this paper, no object mask is provided as initialization during the test phase.
%no knowledge about the objects is provided in the test phase.
Unsupervised segmentation has been addressed by several variants of CNN-based models, such as the two-stream architecture \cite{khoreva2016learning,caelles2017one}, recurrent neural networks \cite{Song_2018_ECCV, tokmakov2017learning} and multi-scale feature fusion \cite{tokmakov2016learning, Song_2018_ECCV}. These approaches generally perform much better than traditional clustering-based pipelines \cite{Chang2013A}. The core idea behind such approaches involves leveraging motion cues explicitly (via optical flow) using a two-steam network \cite{jain2017fusionseg,tokmakov2016learning,tokmakov2017learning}, and/or employing a memory module to capture the evolution of object appearance over time \cite{Song_2018_ECCV,tokmakov2017learning}.

\textbf{Action segmentation.} Action segmentation provides pixel-level localization for actions (\ie action segmentation maps), which are more accurate than bounding boxes for action localization. Lu \etal \cite{lu2015human} propose supervoxel hierarchy to enforce the consistency of the human segmentation in video. Gavrilyuk \etal \cite{gavrilyuk2018actor} infer pixel-level segmentation of an actor and its action in video from a
natural language input sentence.

While we take inspiration from these works, we are the first to present a 3D CNN based deep framework for video object segmentation in a fully automatic manner. Moreover, our proposed method is designed with computational efficiency in mind to enable practical applications for video segmentation.

%\RS{Consider add refs to I3D [Carreira, Zisserman) and S3D [Xie et al. ECCV'18] because S3D also proposes separable 3D convolutions -- for video classification not segmentation}

\section{Generalizing CNN for Dense Prediction from 2D to 3D}
\label{sec:method}
Generalizing CNN framework from  images (2D) to videos (spatio-temporal 3D) involves more work than simply adding one more dimension. A key challenge is due to the asymmetry between space and time. To address the change in apparent size of an object due to perspective, image pipelines crop and reshape images to a fixed size. One cannot do the same with videos since  input videos and the duration of an object track or action in an untrimmed video can vary widely in the temporal dimension. Since an entire video cannot be rescaled to  a fixed size, approaches typically process video as sequences of short (\eg 8-frame) clips.
%Different from images which can be cropped and reshaped into a fixed size, videos vary widely in temporal dimension. To fit input videos into a standard CNN architecture, the videos are divided into fixed length (8 frames) clips. Moreover, there are other challenges such as high computation due to 3D convolution.
In this section, we delve into the details of our network design to address these challenges.

%\subsection{3D feature extractor}
%Feature extractor is a key component in the bottom up module, which is able to obtain higher level representation through selecting and transforming lower level details. To better capture the spatio-temporal information in video, we exploit 3D CNN for action proposal generation and action recognition. One advantage of 3D CNN over 2D CNN is that it encodes motion information by applying convolution in both time and space. Since 3D convolution and 3D max pooling are utilized in the spatial dimension as well as the temporal dimension, the size of video clip is reduced while relevant information is captured. 

\subsection{3D separable convolution with dilation}
\label{sec:3d_separable_conv}
%In T-CNN \cite{hou2017tube}, we show the power of 3D Convolutional Neural Network (3D CNN) for video action detection problem. A natural extension of 3D CNN is to solve video pixel prediction task. 
%However, there are two issues limiting its performance on segmentation. On one hand, with the same input clip, pixel-level prediction task requires more computation resources than the frame-level one since its output dimension is related to its input. On the other hand, 
\begin{figure}
\centering
\includegraphics[width=0.5\linewidth]{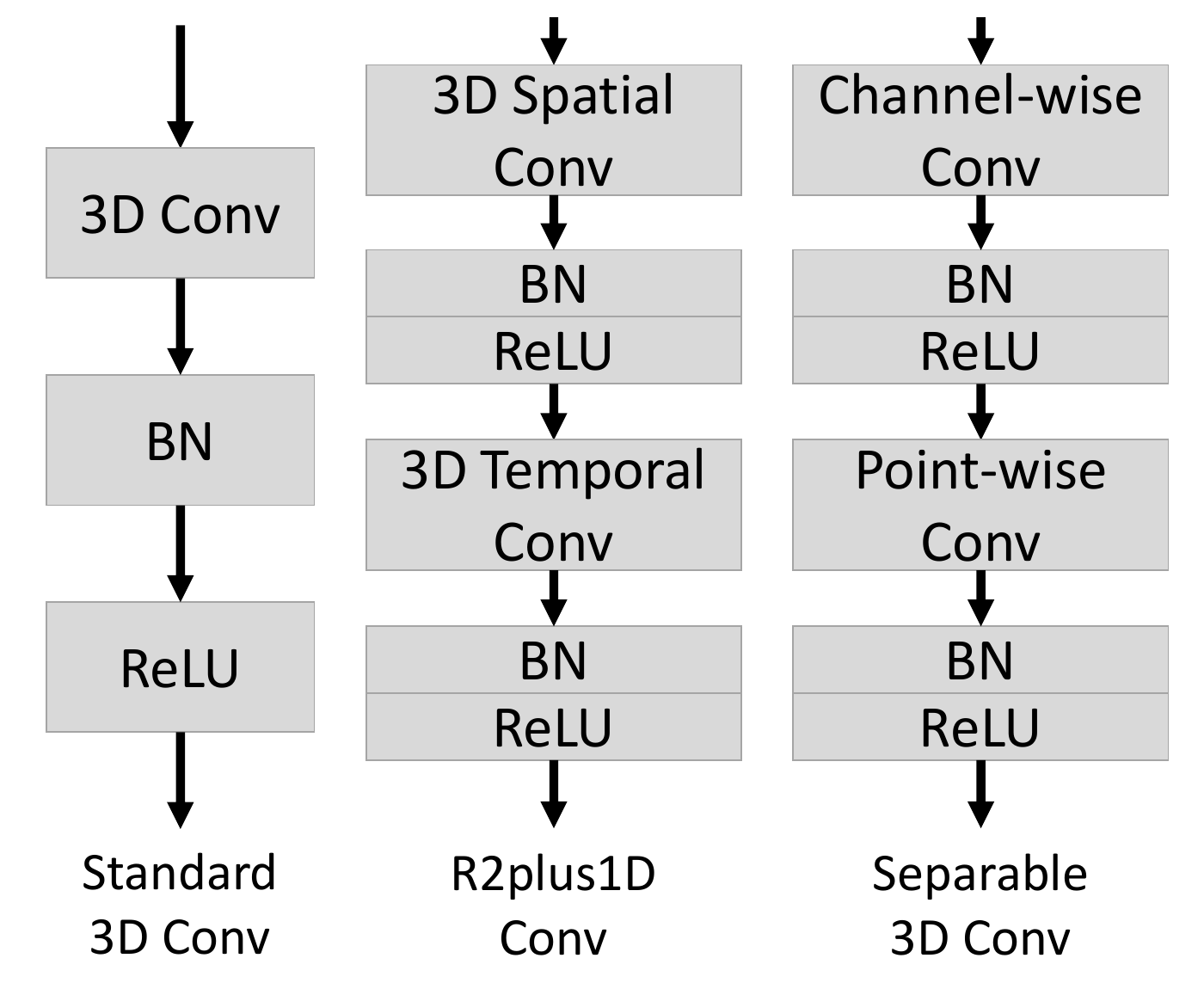}
\caption{Comparison between standard 3D convolution, R2plus1D and 3D separable convolution. R2plus1D factorizes 3D convolution into spatial and temporal convolutions. %By inserting batch normalization and activation layer, R2plus1D has more non-linearity than the standard convolution . 
3D separable convolution is composed of two convolution modules -- channel-wise convolution and point-wise convolution. 
%It is able to save computational cost and provides extra non-linearity. 
}
\label{fig:3d_conv_compare}
\end{figure}
3D CNN training is much less efficient than its 2D counterpart, since 3D kernels and feature maps have more parameters. To address pixel-level prediction in video more efficiently, we propose to use 3D separable convolution in lieu of the standard 3D convolution.
3D separable convolution factorizes a standard 3D convolution into a channel-wise 3D convolution and a point-wise 3D convolution with $1\times1\times1$ kernel. The channel-wise convolution applies a specific filter to each input channel, then point-wise convolution combines the outputs of the channel-wise convolution via a $1\times1\times1$ convolution. This factorization drastically reduces computation and model size.

A standard 3D convolutional operator takes a $H\times W\times T\times M$ feature map $F$ as input and produces a $H\times W\times T\times N$ feature map $G$, where $H$, $W$ and $T$ are the height, width and number of frames (duration) in a clip, respectively. $M$ and $N$ are the number of input/output channels. 
%The $5$-D \MS {why this is 5D? It should be 4D, x, y, t and channel M; you will N of those filters, but convolution is still 4D}  kernel $K$ of 
%A standard 3D convolutional operator has shape $K_h \times K_w \times K_t \times M \times N$, 
 3D convolution kernel has shape $K_h \times K_w \times K_t$, 
where $K_h$, $K_w$ and $K_t$ are the spatial and temporal dimensions of the kernel. 
The computation complexity of a standard 3D convolution is:
\begin{equation}
\label{eq:standard_3d_conv_cost}
H\times W\times T\times M\times N\times K_h\times K_w\times K_t.
\end{equation}
On the other hand, 3D separable convolution is a two-step process. First channel-wise convolution generates the intermediate feature map by applying a specific filter to each input channel ($M$ channels in total).  %Channel-wise convolution can be expressed as:
% \begin{equation}
% \hat{G}_{h,w,t,m} = \sum_{i,j,k} \hat{K}_{i,j,k,m}\cdot F_{h+i-1,w+j-1,t+k-1,m},
% \end{equation}
The computational complexity of channel-wise convolution is:
\begin{equation}
H\times W\times T\times M\times K_h \times K_w\times K_t.
\end{equation}
where the kernel of channel-wise convolution has size $K_h \times K_w \times K_t$. Then point-wise convolution projects intermediate feature map to the final output with computational complexity:
\begin{equation}
H\times W \times T \times M\times N.
\end{equation}

The computational cost of the standard 3D convolution is the multiplication of the feature map dimension ($H, W, T$), input channel dimension $M$, output channel dimension $N$ and kernel dimension ($K_h, K_w, K_t$); while in 3D separable convolution, channel-wise convolution is irrelevant to output filters and point-wise convolution is isolated from kernel dimension. 

Therefore, the computational cost reduction of 3D separable convolution is:
\begin{equation}
\small
\frac{H\times W \times T\times M\times K_h\times K_w\times K_t + H\times W \times T\times M\times N}{H\times W\times T\times M\times N\times K_h\times K_w\times K_t} = \frac{1}{N} + \frac{1}{K_h\times K_w\times K_t}.
\end{equation}

For example, a 3D separable convolution with $3\times3\times3$ kernel dimension and $512$ input/output size only needs about $1/25$ of computational resource compared to that of a standard 3D convolution, leading to significant inference computation reduction.

Meanwhile, in R2plus1D convolution, the standard 3D convolution is factorized into spacial and temporal convolutions. The input feature map with $M$ channels passes through spatial convolution with kernel shape $K_h\times K_w \times 1$ and generates a $M'$ channels intermediate feature map. Then, the intermediate result goes through the temporal convolution with kernel shape $1\times 1\times K_t$ and generates the final output. Compared to standard 3D convolution, the computational cost reduction of R2plus1D convolution is:
\begin{equation}
\label{eqa:r2p1d}
\small
\frac{H\times W \times T\times M\times M'\times K_h\times K_w + H\times W \times T\times M'\times N\times K_t}{H\times W\times T\times M\times N\times K_h\times K_w\times K_t} = \frac{M'}{N\times K_t} + \frac{M'}{M\times K_h\times K_w}.
\end{equation}
The computational reduction is depends on the number of intermediate filters $M'$. According to \cite{r2plus1d_cvpr18}, the number of intermediate filters is set as $\frac{M\times N \times K_t\times K_h\times K_w}{N\times K_t + M\times K_h\times K_w}$ to keep the number of parameters the same as standard convolution.

\textbf{Dilated convolution.} Dilation rate $\gamma_a$ is an attribute of convolution operation. It has the ability to increase the receptive field size, while maintaining the computational cost by skipping $\gamma_a - 1$ entities per valid one in one of dimensions ($x$, $y$, $t$). Adding dilation rates helps capture multi-scale feature representations. Specifically in 3D separable convolution, the dilation rate is only applied to channel-wise convolution, since kernel size of point-wise convolution is fixed at $1\times1\times1$. Moreover, due to the asymmetry between space and time in video, dilation rate in 3D convolution can be divided into two parts -- spatial rate $\gamma_s$ and temporal rate $\gamma_t$. By using the dilation rate, the channel-wise convolution can be expressed as follows:
\begin{equation}
\label{eq:dilated_conv}
\hat{G}_{h,w,t,m} = \sum_{i,j,k} \hat{K}_{i,j,k,m}\cdot F_{h+\gamma_s\cdot i,w+\gamma_s\cdot j,t+\gamma_t\cdot k,m}.
\end{equation}
As shown in Eq.~\ref{eq:dilated_conv}, the computational cost does not change by adding the dilation rate.

\subsection{Network architecture}
\label{sec:network}

\begin{figure}[!t]
\centering
\includegraphics[width=0.95\linewidth]{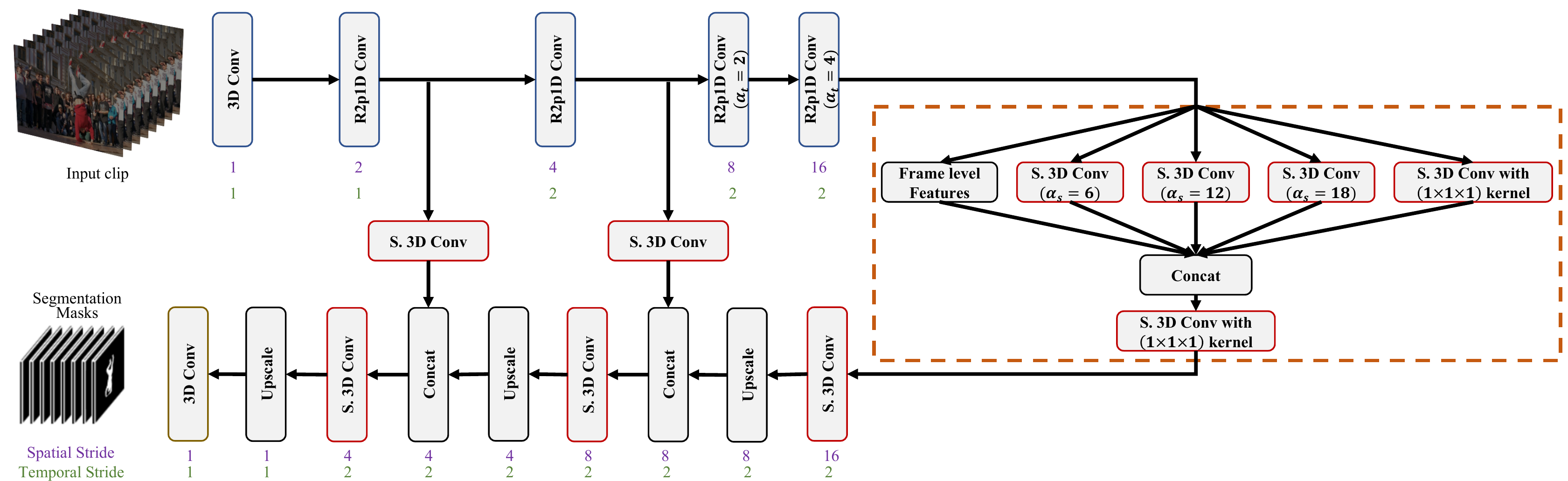}
\caption{The network architecture of our method for video object segmentation. It has three components: an encoder (feature extractor), a pyramid pooling module and a decoder to recover the spatial-temporal details gradually. The encoder, shown in the top-left, takes pixel values as input, extracts features layer by layer, and generates rich contextual feature as the final output. Moreover, its intermediate results are merged with the decoder module. The pyramid pooling module connects the encoder and decoder modules. It varies the receptive field size by modifying the spatial stride. ``S. 3D Conv'' indicates the 3D separable convolution. ``Frame level features'' block is composed by a reduced average pooling and duplication to generate the frame level features with the same dimension as the other branches.
%The decoder module is able to recover the details and output the pixel-level output. \MS{Rui where is the classification branch here?}
}
\label{fig:segmentation}
\end{figure}

The proposed 3D CNN architecture for action/object segmentation in video is illustrated in Figure~\ref{fig:segmentation}.
The network builds upon an encoder-decoder structure for image semantic segmentation. 
%\MS{We use the same network for video object segmentation and video action segmentation. In the first case, we assign each pixel object or background labels. In the second case, given $k$ total number of actions,   we assign each pixel one of $k$ action labels or background label.  Here we will give details about network taking  video object segmentation as an example.} 
A video is divided into 8-frame clips as input to the network. In the encoder module, 3D convolution are performed. To capture higher level information, the spatial and temporal sizes are reduced. In order to generate the pixel-wise segmentation map for each frame in the original size, 3D upsampling is used in the decoder module to increase the resolution of feature maps. To capture spatial and temporal information at different scales, a concatenation with the corresponding feature maps from the encoder module is employed after each 3D upsampling layer. Finally, a softmax cross entropy loss layer is used for pixel-wise prediction (\ie~background or object foreground) for each frame in a clip.

In the encoder, we adopt R2plus1D~\cite{r2plus1d_cvpr18} as feature extractor to leverage its pre-trained model on large scale action recognition dataset. We modify the last three groups of convolutional layers by adjusting the dilation rate to keep temporal stride as $4$.
We also insert a 3D pyramid pooling before the decoder. 
In the decoder, the up-sampling layer group is composed of a tri-linear interpolation layer, a 3D separable convolution layer as well as a feature concatenation layer incorporating encoder feature.

\textbf{3D Pyramid Pooling.}
Compared to the 2D counterpart \cite{chen2018deeplab}, pyramid pooling in 3D is more challenging. In spatial domain, multi-scale information should be captured. In temporal domain, detailed motion information should be preserved as well.
Therefore, our 3D pyramid pooling has 5 branches. The details are listed as follows:
\begin{enumerate}%[topsep=0pt,itemsep=2pt,parsep=2pt]
\item 3D separable convolutions with kernel size $3\times3\times3$ and different spatial dilation rates $\gamma_s = {6, 12, 18}$. The convolutions with multiple spatial dilation rates are able to capture multi-scale information. 

\item 3D separable convolution with kernel size $1\times1\times1$ -- a default layer for dense prediction.

\item There are two steps to get the frame feature. First, input is average pooled with kernel $H\times W \times 1$. For the average pooling output, its spatial dimension is $1$, while its temporal length and filter size are the same as those of the input. Then the average pooling output is upscaled to the input size.
\end{enumerate}

By concatenating the $5$ branches together, and applying a $1\times1\times1$ convolution, the final output is obtained and fed into the decoder module.

\section{Experiments}
\label{sec:experiments}
\textbf{Implementation details.} 
To verify the effectiveness of the proposed 3D CNN framework for action/object segmentation in videos, we evaluate our approach on two video object segmentation datasets -- DAVIS'16 \cite{Perazzi2016} and Segtrack-v2 \cite{li2013video}, and a video action segmentation dataset -- J-HMDB~\cite{Jhuang:ICCV:2013}. We adopt R2plus1d encoder pretrained on Kinetics dataset~\cite{zisserman2017kinetics} and leverage the Adam optimizer with initial learning rate $10^{-4}$, exponential decay rate $0.95$ and decay step $1$ epoch. The model is trained for $100$ epochs with exponential decay.
In 3D spatial pyramid pooling module, a large crop size is required to ensure the convolution kernel with spatial dilation rate is effective; otherwise, the filter weights with large dilation rate are mostly applied to the padded zero region. Therefore, We employ a spatial crop size of $384$ during both training and testing.
During training, we randomly scale the resolution of input clips with ratio $[0.5,2]$ and apply a random horizontal flip.

\textbf{Experiments on DAVIS'16.}
Densely Annotated Video Segmentation 2016 (DAVIS'16) dataset is a benchmark dataset for video object segmentation. It consists of 50 videos with 3455 annotated frames. Consistent with most prior work, we conduct experiments on the 480p videos with a resolution of $854 \times 480$ pixels. 30 videos are used for training and 20 for validation.
We adopt the same evaluation setting in \cite{Perazzi2016}. There are three parts. \textbf{Region Similarity} $\mathcal{J}$, which is obtained by IoU (Intersection over Union) between the prediction and the ground-truth segmentation map. \textbf{Contour Accuracy} $\mathcal{F}$ measures the contours accuracy. \textbf{Temporal Stability} $\mathcal{T}$ tracks the temporal consistency in a video. For the first two evaluation, we report the mean, recall and decay. For the third one, we report the average.

\begin{figure}[!htbp]
\centering
\includegraphics[width=0.95\linewidth]{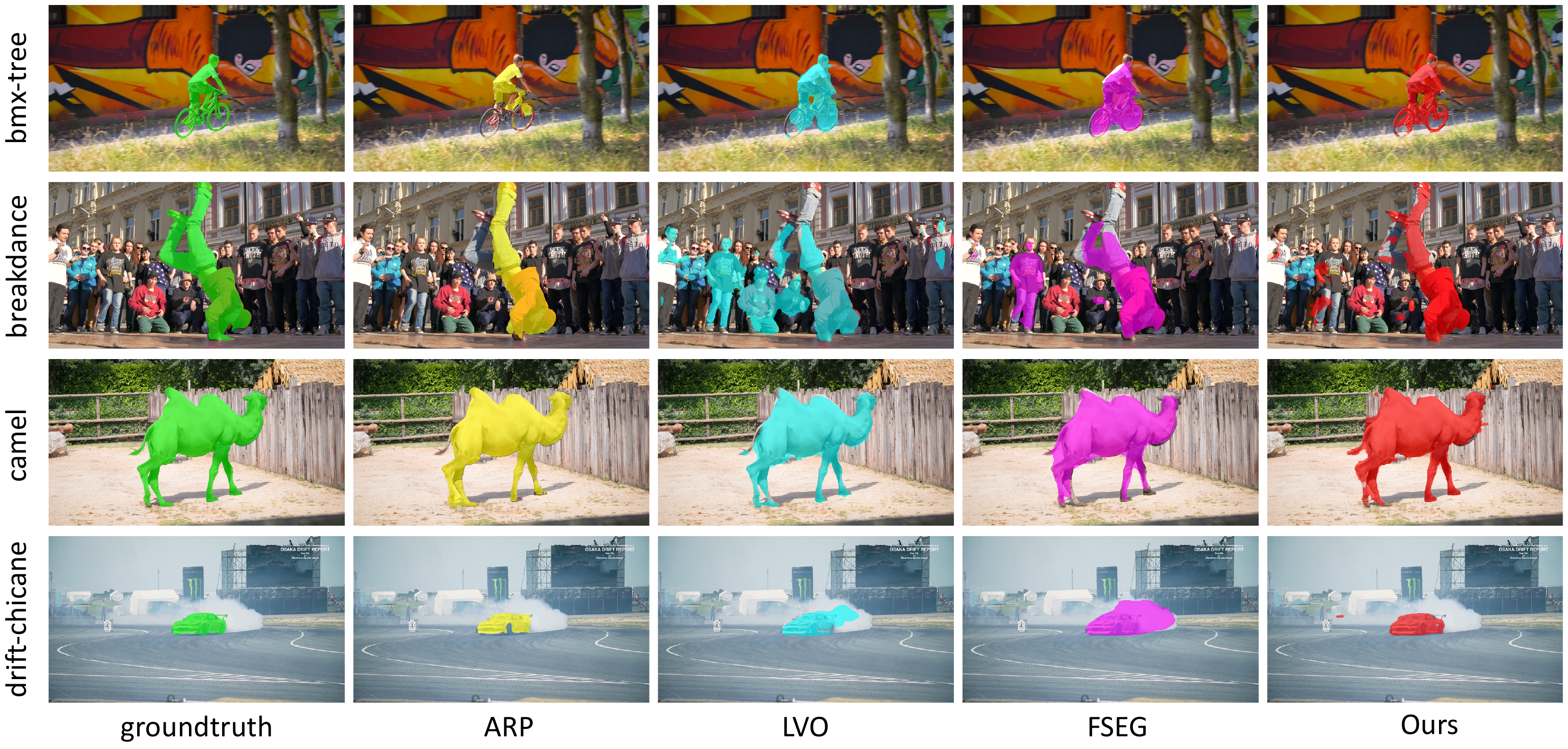}
\caption{Qualitative results of the proposed (Ours) approach (red), ARP (yellow), LVO (cyan) and FSEG (magenta) on selected frames from DAVIS dataset.}
\label{fig:mtcnn_example}
\end{figure}

\begin{table}[!hbp]

\centering
\footnotesize
\begin{tabular}{c|c|cccccccccc}%p{0.5cm}p{0.5cm}p{0.44cm}p{0.5cm}p{0.5cm}p{0.5cm}p{0.5cm}p{0.5cm}p{0.5cm}p{0.5cm}p{0.5cm}}

\hline

\multicolumn{2}{c|}{\multirow{2}{*}{Measure}}    & MotAdapt  & LSMO  & PDB   & ARP  & FSEG   & LMP & FST    & CUT  & NLC & Ours \\
\multicolumn{2}{c|}{}           & \cite{siam2018video}  & \cite{tokmakov2019learning} & \cite{Song_2018_ECCV} & \cite{koh2017primary} & \cite{jain2017fusionseg}  & \cite{tokmakov2016learning}  & \cite{keuper2015motion}   & \cite{faktor2014video}  & \cite{fragkiadaki2012video} & \\ 
\hline
\multirow{3}{*}{$\mathcal{J}$}  
& Mean $\uparrow$               & 77.2  & 78.2  & 77.2  & 76.2	& 70.7	
                                & 70.0	& 55.8	& 55.2	& 55.1  & \textbf{78.3}  \\
& Recall $\uparrow$             & 87.8  & 89.1  & 90.1  & \textbf{91.1}	& 83.5	
                                & 85.0	& 64.9	& 57.5  & 55.8  & \textbf{91.1} \\
& Decay $\downarrow$            & 5.0   & 4.1   & 0.9   & 7.0	& 1.5	
                                & 1.3	& \textbf{0.0}	& 2.2   & 12.6  & 2.3 \\
\hline
\multirow{3}{*}{$\mathcal{F}$}  
& Mean $\uparrow$               & \textbf{77.4}  & 75.9  & 74.5  & 70.6	& 65.3	
                                & 65.9	& 51.1	& 55.2	& 52.3  & 77.2 \\
& Recall $\uparrow$             & 84.4  & \textbf{84.7}  & 84.4  & 83.5  & 73.8	
                                & 79.2	& 51.6	& 61.0	& 51.9  & \textbf{84.7} \\
& Decay $\downarrow$            & 3.3   & 3.5   & \textbf{-0.2}  & 7.9   & 1.8	
                                & 2.5	& 2.9	& 3.4	& 11.4  & 4.9 \\
\hline
$\mathcal{T}$
& Mean $\downarrow$             & 27.9  & \textbf{21.2}  & 29.1  & 39.3	& 32.8	
                                & 57.2	& 36.6	& 27.7	& 42.5  & 22.0\\
\hline
\end{tabular}
\caption{Overall results of region similarity ($\mathcal{J}$), contour accuracy ($\mathcal{F}$) and temporal stability ($\mathcal{T}$) for different approaches. $\uparrow$ means the higher the better, and $\downarrow$ means the lower the better.}
\label{tab:davis_overall}
\end{table}
We compare our results with several unsupervised approaches, since our approach does not require any manual annotation or prior information about the object to be segmented.
%, which is defined as unsupervised segmentation.
We cannot compare directly to semi-supervised approaches that require the ground truth segmentation map in the first frame of each test video to be given. Table~\ref{tab:davis_overall} summarizes the performance of our method against the state-of-the-art unsupervised approaches on DAVIS'16. Our method achieves the best performance in all performance metrics. Compared to ARP \cite{koh2017primary}, the previous state-of-the-art unsupervised approach, our method achieves 5\% gain in contour accuracy ($\mathcal{F}$) and 15\% gain in temporal stability ($\mathcal{T}$), demonstrating that 3D CNN can effectively take advantage of the temporal information in video frames to achieve temporal segmentation consistency. We also present the qualitative results on four video sequences in Figure~\ref{fig:mtcnn_example}.

\textbf{Experiments on Segtrack-v2.} SegTrack-v2 \cite{li2013video} contains $14$ video clips with 24 objects and 947 frames. Pixel-level object mask on each frame is provided. We adopt the same evaluation setting in \cite{khoreva2016learning} and report the mean Intersection over Union. For videos with multiple instances with individual ground-truth segmentation mask, we group them as a single foreground for evaluation. Compared with other unsupervised video object segmentation methods, our proposed approach outperforms all of them (see table \ref {tab:segtrack}).
\begin{table}[]
\centering
\small
\begin{tabular}{l|ccccc}
\hline
Method      & FST \cite{papazoglou2013fast}  & KEY \cite{lee2011key}  & LSMO \cite{tokmakov2017learning} & FSEG \cite{jain2017fusionseg} & Ours \\
\hline
mean IoU    & 53.5  & 57.3  & 53.7  & 61.4  & {\bf 62.1} \\   
\hline
\end{tabular}
\caption{Video object segmentation results on Segtrack-v2 dataset. We compare the performance of our approach with other state-of-the-art unsupervised approaches.}
\label{tab:segtrack}
\end{table}

\textbf{Experiments on J-HMDB.}
The J-HMDB dataset consists of 928 videos with 21 different actions. There are three train-test splits and the evaluation is done on the average results over the three splits. We leverage the mask annotation provided from J-HMDB dataset and train our semantic segmentation pipeline to segment the foreground action. Then the region with maximum area is selected through connected component \cite{he2009fast}. On the final feature map of each frame, we crop a box to tightly surround the selected foreground region and resize the box into a fixed shape. Finally, the cropped feature map goes through a classifier to predict action classes. We use softmax cross entropy loss to back propagate gradients for all the weights. For evaluation, we followed the metrics in \cite{gavrilyuk2018actor}. Mean IoU is computed as the average over the IoU of each test sample. In addition, frame-level mean Average Precision (mAP) is evaluated as well. Since bounding boxes detection can be obtained by selecting the tightest rectangle region enclosing the segmentation mask, we are also able to compare with the state-of-the-art action detection approaches. 
\begin{table}[!htbp]
\centering
\small
\begin{tabular}{lccc}
\hline
                                                & mean IoU      & frame-mAP ($\alpha=0.5$)\\
\hline
Lu \etal~\cite{lu2015human}                     & 48.8          & -- \\
Gavrilyuk \etal~\cite{gavrilyuk2018actor}       & 54.2          & -- \\
\hline
Kalogeiton \etal \cite{kalogeiton2017action}    & --            & 65.7 \\
Duarte \etal \cite{duarte2018videocapsulenet}   & --            & 64.6 \\
Hou \etal \cite{hou2017tube}                    & --            & 61.3 \\
\hline
\textbf{Ours}                                   & \textbf{68.1} & \textbf{68.9} \\
\hline
\end{tabular}
\caption{Comparison with the state-of-the-art action segmentation and detection approaches on J-HMDB. The first part of the table shows the mean Intersection-over-Union (IoU) of action segmentation approaches and the second part shows the frame level mean Average Precision (mAP) of CNN based action detection approaches.}
\label{tab:jhmdb-segmentation}
\end{table}

\begin{figure}[!t]
\centering
\includegraphics[width=0.98\linewidth]{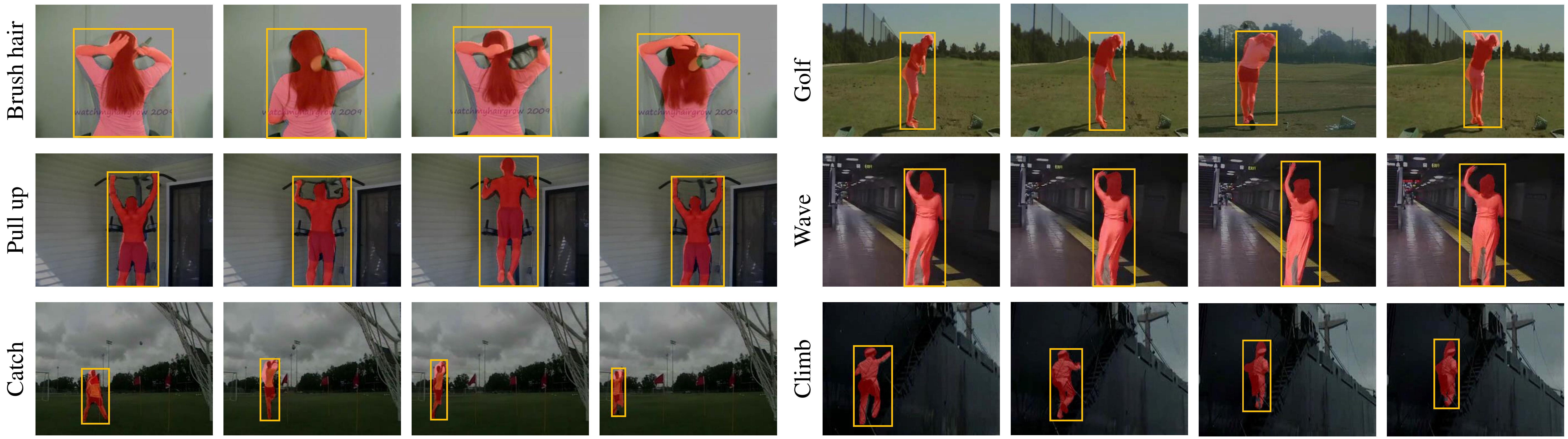}
\caption{Action segmentation and detection results obtained by our method on the J-HMDB dataset. Red pixel-wise segmentation maps show the predictions, and the yellow boxes show the bounding boxes generated from the segmentation maps.}
\label{fig:jhmdb_seg_results}
\end{figure}

Table~\ref{tab:jhmdb-segmentation} reports the action segmentation/detection results of our method and the state-of-the-art approaches. It is evident that our method outperforms these methods considerably in evaluation metrics. Figure~\ref{fig:jhmdb_seg_results} presents both action segmentation (in red) and bounding boxes detection (in yellow) results on several video sequences from J-HMDB.

\subsection{Ablation Study}
To better understand the contribution of each component in our proposed approach, we conduct video object segmentation on DAVIS'16 with different settings, summarized in Table~\ref{tab:ablation} and Table~\ref{tab:var_convs}.

%\textbf{Effectiveness of 3D proposal}.
\textbf{Dilation rate.} We adopt R2Plus1D pre-trained model in our approach, and increase the feature map's temporal size of last two convolution layer group by specifying the dilation rate. 
As shown in first section of Table~\ref{tab:ablation}, the performance is boosted by increasing the number of frames in the final feature map. However, too large feature map (8 frames) will cause out of memory error, that is the reason we stop at 4 frames.

% Let $\gamma_{t_i}$ be the output stride of $i$-th convolutional layer and $\gamma_{t}$ is the designed output stride of the final feature map, where 
% \[
%     \gamma_{t_i}= 
% \begin{cases}
%     \gamma_t_,& \text{if } \Gamma_t < \gamma'_{t_i}\\
%     \gamma'_{t_i},              & \text{otherwise}
% \end{cases}
% \]
% The temporal dilation rate is defined by $d_{t_i} = \gamma'_{t_i} / \gamma_{t_i}$. By increasing the bottom-up module duration, the segmentation performance is boosted from $62.4$ to $68.1$ and $72.4$ by double and quadruple the duration of bottom-up module output.
%However, continued increasing the temporal size is not sufficient, since larger feature map needs extra storage and computation resources.

\textbf{3D pyramid pooling.} As shown in the second section of Table~\ref{tab:ablation}, inserting 3D pyramid pooling module improves the segmentation accuracy by over 5\%. We experiment multiple branches with different spatial dilation rates (as noted in the bracket). According to the experimental results, including more branches with larger receptive field ($4$ branches) or larger temporal stride is not helpful. When the receptive field size is close to or larger than the feature map size, most of the filters cannot capture any useful information, since they only cover padded zeros instead of valid area. Segmentation accuracy further improves when frame-level features are added.

\begin{table}[!htbp]
\centering
\begin{tabular}{c|c|c}
\hline
Final feature map dim.  & 3D Pyramid Pooling    & $\mathcal{J}$-Mean \\
%Temporal Rate $(\gamma_{t4}, \gamma_{t5})$
\hline
$1\times20\times20$     & --                    & 64.5 \\
$2\times20\times20$     & --                    & 69.1 \\
$4\times20\times20$     & --                    & 71.3 \\
\hline
$4\times20\times20$     & (6, 12, 18)           & 78.1 \\
$4\times20\times20$     & (6, 12, 18) + FF      & 78.3 \\
$4\times20\times20$     & (6, 12, 18, 24)       & 77.9 \\
$4\times20\times20$     & (8, 16, 24)           & 74.9 \\
\hline
\end{tabular}
\caption{Ablation study of our method for video object segmentation on DAVIS-16. 
In the first horizontal section of the Table, we investigate various temporal size of the final feature map by increasing the temporal dilation rate in the last two convolutional layer groups. In the second section of the Table, we fix feature map dimensions (including temporal dimension) and explore different settings in 3D pyramid pooling module (shown in the second column), which includes various receptive fields and whether frame-level features (FF) are included or not. Specifically, the numbers in the parentheses () indicate spatial dilation rates of branches. We compare  performance with different sizes of feature maps,  
%(left of bold vertical line), 
with or without 3D pyramid pooling layer. 
%(right of bold vertical line).
Mean IoU is used as evaluation metric as shown in third column. 
}
\label{tab:ablation}
\end{table}

\textbf{3D separable convolution}. The proposed pipeline leverages 3D separable convolution in pyramid pooling and decoder. We perform an experiment by replacing all the 3D separable convolutions with R2plus1D convolutions and standard 3D convolutions.
The comparison of performance and computational cost is shown in Table~\ref{tab:var_convs}. All the experiments are carried out on a workstation with a NVIDIA Titan XP GPU and PyTorch. We only take 3D pyramid pooling and decoder part during inference into computation cost, since all of them share the same pre-trained model. ``Operations'' counts the total number of additions and multiplications. ``GPU Mem.'' is the size of allocated GPU memory. With the same input (a $8$ frames clip with resolution $320\times320$), 3D separable convolution greatly reduces the computational cost without sacrificing the performance.
%while keeping the competitive performance.

\begin{table}[!htbp]
\centering
\begin{tabular}{l|ccc}
\hline
Conv. Type          & Operations    & GPU Mem.  & $\mathcal{J}$-Mean \\
\hline
Standard 3D Conv.   & 136 Billion   & 255 MB    & 77.1\\
R2plus1D Conv.      & 136 Billion   & 256 MB    & 77.6\\
3D Separable Conv.  & 6 Billion     & 11 MB     & 77.4\\
\hline
\end{tabular}
\caption{The comparison of performance and computational cost of different 3D convolutions. 3D separable convolution is able to reach the similar accuracy with only 5\% of parameters. In R2plus1D, we adopt the settings in \cite{r2plus1d_cvpr18}, which sets the number of intermediate filters to be  the same as the number of parameters of the standard 3D convolution.} % Standard 3D Conv.}
\label{tab:var_convs}
\end{table}

\section{Conclusion}
\label{sec:conclusion}
This paper proposes a fully convolutional 3D CNN pipeline for action/object segmentation in video. The approach leverages a model pre-trained on large-scale action recognition task as an encoder to enable us to perform unsupervised video object segmentation (\ie generate pixel-level object masks without initialization). We also use separable filters to significantly reduce the computational burden of the standard 3D convolutions. Extensive experiments on several benchmark datasets demonstrate the strength of our approach for spatio-temporal action segmentation as well as video object segmentation compared with the state-of-the-art approaches.

%\vspace{0.75em}
%\noindent\textbf{Acknowledgement.}
\section{Acknowledgements}
This research is based upon work supported in part by the National Science Foundation under Grants No.\ 1741431 and the Office of the Director of National Intelligence (ODNI), Intelligence Advanced Research Projects Activity (IARPA), via IARPA R\&D Contract No.\ D17PC00345. The views, findings, opinions, and conclusions or recommendations contained herein are those of the authors and should not be interpreted as necessarily representing the official policies or endorsements, either expressed or implied, of the NSF, ODNI, IARPA, or the U.S.\ Government. The U.S.\ Government is authorized to reproduce and distribute reprints for Governmental purposes notwithstanding any copyright annotation thereon.

\bibliographystyle{abbrv}
\bibliography{egbib}

\begin{thebibliography}{48}
\providecommand{\natexlab}[1]{#1}
\providecommand{\url}[1]{\texttt{#1}}
\expandafter\ifx\csname urlstyle\endcsname\relax
  \providecommand{\doi}[1]{doi: #1}\else
  \providecommand{\doi}{doi: \begingroup \urlstyle{rm}\Url}\fi

\bibitem[Badrinarayanan et~al.(2017)Badrinarayanan, Kendall, and
  Cipolla]{badrinarayanan2015segnet}
Vijay Badrinarayanan, Alex Kendall, and Roberto Cipolla.
\newblock Segnet: A deep convolutional encoder-decoder architecture for image
  segmentation.
\newblock \emph{IEEE transactions on pattern analysis and machine
  intelligence}, 39\penalty0 (12):\penalty0 2481--2495, 2017.

\bibitem[Caelles et~al.(2017)Caelles, Maninis, Pont-Tuset, Leal-Taix{\'e},
  Cremers, and Van~Gool]{caelles2017one}
Sergi Caelles, Kevis-Kokitsi Maninis, Jordi Pont-Tuset, Laura Leal-Taix{\'e},
  Daniel Cremers, and Luc Van~Gool.
\newblock One-shot video object segmentation.
\newblock In \emph{IEEE Conference on Computer Vision and Pattern Recognition},
  2017.

\bibitem[Carreira and Zisserman(2017)]{carreira2017quo}
Joao Carreira and Andrew Zisserman.
\newblock Quo vadis, action recognition? a new model and the {Kinetics}
  dataset.
\newblock In \emph{proceedings of the IEEE Conference on Computer Vision and
  Pattern Recognition}, pages 6299--6308, 2017.

\bibitem[Chang et~al.(2013)Chang, Wei, and {III}]{Chang2013A}
Jason Chang, Donglai Wei, and John W.~Fisher {III}.
\newblock A video representation using temporal superpixels.
\newblock In \emph{IEEE Conference on Computer Vision Pattern Recognition},
  2013.

\bibitem[Chen et~al.(2017)Chen, Papandreou, Schroff, and
  Adam]{chen2017rethinking}
Liang-Chieh Chen, George Papandreou, Florian Schroff, and Hartwig Adam.
\newblock Rethinking atrous convolution for semantic image segmentation.
\newblock \emph{arXiv preprint arXiv:1706.05587}, 2017.

\bibitem[Chen et~al.(2018)Chen, Papandreou, Kokkinos, Murphy, and
  Yuille]{chen2018deeplab}
Liang-Chieh Chen, George Papandreou, Iasonas Kokkinos, Kevin Murphy, and Alan~L
  Yuille.
\newblock {DeepLab}: Semantic image segmentation with deep convolutional nets,
  atrous convolution, and fully connected {CRFs}.
\newblock \emph{IEEE transactions on pattern analysis and machine
  intelligence}, 40\penalty0 (4):\penalty0 834--848, 2018.

\bibitem[Duarte et~al.(2018)Duarte, Rawat, and Shah]{duarte2018videocapsulenet}
Kevin Duarte, Yogesh Rawat, and Mubarak Shah.
\newblock Videocapsulenet: A simplified network for action detection.
\newblock In \emph{Advances in Neural Information Processing Systems}, pages
  7610--7619, 2018.

\bibitem[Dutt~Jain et~al.(2017)Dutt~Jain, Xiong, and
  Grauman]{jain2017fusionseg}
Suyog Dutt~Jain, Bo~Xiong, and Kristen Grauman.
\newblock {FusionSeg}: Learning to combine motion and appearance for fully
  automatic segmentation of generic objects in videos.
\newblock In \emph{Proceedings of the IEEE Conference on Computer Vision and
  Pattern Recognition}, pages 3664--3673, 2017.

\bibitem[Faktor and Irani(2014)]{faktor2014video}
Alon Faktor and Michal Irani.
\newblock Video segmentation by non-local consensus voting.
\newblock In \emph{British Machine and Vision Conference}, volume~2, 2014.

\bibitem[Fragkiadaki et~al.(2012)Fragkiadaki, Zhang, and
  Shi]{fragkiadaki2012video}
Katerina Fragkiadaki, Geng Zhang, and Jianbo Shi.
\newblock Video segmentation by tracing discontinuities in a trajectory
  embedding.
\newblock In \emph{IEEE Conference on Computer Vision and Pattern Recognition},
  2012.

\bibitem[Gavrilyuk et~al.(2018)Gavrilyuk, Ghodrati, Li, and
  Snoek]{gavrilyuk2018actor}
Kirill Gavrilyuk, Amir Ghodrati, Zhenyang Li, and Cees~GM Snoek.
\newblock Actor and action video segmentation from a sentence.
\newblock In \emph{IEEE Conference on Computer Vision and Pattern Recognition},
  2018.

\bibitem[Hara et~al.(2018)Hara, Kataoka, and Satoh]{hara2018can}
Kensho Hara, Hirokatsu Kataoka, and Yutaka Satoh.
\newblock Can spatiotemporal 3d {CNN}s retrace the history of 2d {CNN}s and
  {ImageNet}?
\newblock In \emph{Proceedings of the IEEE conference on Computer Vision and
  Pattern Recognition}, pages 6546--6555, 2018.

\bibitem[He et~al.(2016)He, Zhang, Ren, and Sun]{he2016deep}
Kaiming He, Xiangyu Zhang, Shaoqing Ren, and Jian Sun.
\newblock Deep residual learning for image recognition.
\newblock In \emph{Proceedings of the IEEE conference on computer vision and
  pattern recognition}, pages 770--778, 2016.

\bibitem[He et~al.(2009)He, Chao, Suzuki, and Wu]{he2009fast}
Lifeng He, Yuyan Chao, Kenji Suzuki, and Kesheng Wu.
\newblock Fast connected-component labeling.
\newblock \emph{Pattern Recognition}, 42\penalty0 (9):\penalty0 1977--1987,
  2009.

\bibitem[Jhuang et~al.(2013)Jhuang, Gall, Zuffi, Schmid, and
  Black]{Jhuang:ICCV:2013}
H.~Jhuang, J.~Gall, S.~Zuffi, C.~Schmid, and M.~J. Black.
\newblock Towards understanding action recognition.
\newblock In \emph{Proceedings of the IEEE International Conference on Computer
  Vision}, 2013.

\bibitem[Kalogeiton et~al.(2017)Kalogeiton, Weinzaepfel, Ferrari, and
  Schmid]{kalogeiton2017action}
Vicky Kalogeiton, Philippe Weinzaepfel, Vittorio Ferrari, and Cordelia Schmid.
\newblock Action tubelet detector for spatio-temporal action localization.
\newblock In \emph{Proceedings of the IEEE International Conference on Computer
  Vision}, 2017.

\bibitem[Karpathy et~al.(2014)Karpathy, Toderici, Shetty, Leung, Sukthankar,
  and Fei-Fei]{karpathy2014large}
Andrej Karpathy, George Toderici, Sanketh Shetty, Thomas Leung, Rahul
  Sukthankar, and Li~Fei-Fei.
\newblock Large-scale video classification with convolutional neural networks.
\newblock In \emph{IEEE Conference on Computer Vision and Pattern Recognition},
  2014.

\bibitem[Kay et~al.(2017)Kay, Carreira, Simonyan, Zhang, Hillier,
  Vijayanarasimhan, Viola, Green, Back, Natsev, et~al.]{zisserman2017kinetics}
Will Kay, Joao Carreira, Karen Simonyan, Brian Zhang, Chloe Hillier, Sudheendra
  Vijayanarasimhan, Fabio Viola, Tim Green, Trevor Back, Paul Natsev, et~al.
\newblock The {Kinetics} human action video dataset.
\newblock \emph{arXiv preprint arXiv:1705.06950}, 2017.

\bibitem[Ke et~al.(2007)Ke, Sukthankar, and Hebert]{ke2007event}
Yan Ke, Rahul Sukthankar, and Martial Hebert.
\newblock Event detection in crowded videos.
\newblock In \emph{Proceedings of the IEEE International Conference on Computer
  Vision}, 2007.

\bibitem[Keuper et~al.(2015)Keuper, Andres, and Brox]{keuper2015motion}
Margret Keuper, Bjoern Andres, and Thomas Brox.
\newblock Motion trajectory segmentation via minimum cost multicuts.
\newblock In \emph{Proceedings of the IEEE International Conference on Computer
  Vision}, 2015.

\bibitem[Koh and Kim(2017)]{koh2017primary}
Yeong~Jun Koh and Chang-Su Kim.
\newblock Primary object segmentation in videos based on region augmentation
  and reduction.
\newblock In \emph{IEEE Conference on Computer Vision and Pattern Recognition},
  2017.

\bibitem[Krizhevsky et~al.(2012)Krizhevsky, Sutskever, and
  Hinton]{krizhevsky2012imagenet}
Alex Krizhevsky, Ilya Sutskever, and Geoffrey~E Hinton.
\newblock {ImageNet} classification with deep convolutional neural networks.
\newblock In \emph{Advances in neural information processing systems}, pages
  1097--1105, 2012.

\bibitem[LeCun et~al.(2015)LeCun, Bengio, and Hinton]{lecun2015deep}
Yann LeCun, Yoshua Bengio, and Geoffrey Hinton.
\newblock Deep learning.
\newblock \emph{Nature}, 521\penalty0 (7553), 2015.

\bibitem[Lee et~al.(2011)Lee, Kim, and Grauman]{lee2011key}
Yong~Jae Lee, Jaechul Kim, and Kristen Grauman.
\newblock Key-segments for video object segmentation.
\newblock In \emph{Proceedings of the IEEE International Conference on Computer
  Vision}, 2011.

\bibitem[Li et~al.(2013)Li, Kim, Humayun, Tsai, and Rehg]{li2013video}
Fuxin Li, Taeyoung Kim, Ahmad Humayun, David Tsai, and James~M Rehg.
\newblock Video segmentation by tracking many figure-ground segments.
\newblock In \emph{Proceedings of the IEEE International Conference on Computer
  Vision}, pages 2192--2199, 2013.

\bibitem[Long et~al.(2015)Long, Shelhamer, and Darrell]{long2015fully}
Jonathan Long, Evan Shelhamer, and Trevor Darrell.
\newblock Fully convolutional networks for semantic segmentation.
\newblock In \emph{Proceedings of the IEEE conference on computer vision and
  pattern recognition}, pages 3431--3440, 2015.

\bibitem[Lu et~al.(2015)Lu, Corso, et~al.]{lu2015human}
Jiasen Lu, Jason~J Corso, et~al.
\newblock Human action segmentation with hierarchical supervoxel consistency.
\newblock In \emph{IEEE Conference on Computer Vision and Pattern Recognition},
  2015.

\bibitem[Maninis et~al.(2018)Maninis, Caelles, Chen, Pont-Tuset, and
  Gool]{Maninis2018Video}
Kevis~Kokitsi Maninis, Sergi Caelles, Yuhua Chen, Jordi Pont-Tuset, and Luc~Van
  Gool.
\newblock Video object segmentation without temporal information.
\newblock \emph{IEEE transactions on pattern analysis and machine
  intelligence}, PP\penalty0 (99):\penalty0 1--1, 2018.

\bibitem[Noh et~al.(2015)Noh, Hong, and Han]{noh2015learning}
Hyeonwoo Noh, Seunghoon Hong, and Bohyung Han.
\newblock Learning deconvolution network for semantic segmentation.
\newblock In \emph{Proceedings of the IEEE international conference on computer
  vision}, pages 1520--1528, 2015.

\bibitem[Papazoglou and Ferrari(2013)]{papazoglou2013fast}
Anestis Papazoglou and Vittorio Ferrari.
\newblock Fast object segmentation in unconstrained video.
\newblock In \emph{Proceedings of the IEEE International Conference on Computer
  Vision}, 2013.

\bibitem[Perazzi et~al.(2016)Perazzi, Pont-Tuset, McWilliams, {Van Gool},
  Gross, and Sorkine-Hornung]{Perazzi2016}
F.~Perazzi, J.~Pont-Tuset, B.~McWilliams, L.~{Van Gool}, M.~Gross, and
  A.~Sorkine-Hornung.
\newblock A benchmark dataset and evaluation methodology for video object
  segmentation.
\newblock In \emph{IEEE Conference on Computer Vision and Pattern Recognition},
  2016.

\bibitem[Perazzi et~al.(2017)Perazzi, Khoreva, Benenson, Schiele, and
  Sorkine-Hornung]{khoreva2016learning}
Federico Perazzi, Anna Khoreva, Rodrigo Benenson, Bernt Schiele, and Alexander
  Sorkine-Hornung.
\newblock Learning video object segmentation from static images.
\newblock In \emph{Proceedings of the IEEE Conference on Computer Vision and
  Pattern Recognition}, pages 2663--2672, 2017.

\bibitem[Ronneberger et~al.(2015)Ronneberger, Fischer, and
  Brox]{ronneberger2015u}
Olaf Ronneberger, Philipp Fischer, and Thomas Brox.
\newblock U-net: Convolutional networks for biomedical image segmentation.
\newblock In \emph{International Conference on Medical Image Computing and
  Computer-Assisted Intervention}, pages 234--241. Springer, 2015.

\bibitem[Rui et~al.(2017)Rui, Chen, and Shah]{hou2017tube}
Hou Rui, Chen Chen, and Mubarak Shah.
\newblock Tube convolutional neural network ({T-CNN}) for action detection in
  videos.
\newblock In \emph{Proceedings of the IEEE International Conference on Computer
  Vision}, 2017.

\bibitem[Siam et~al.(2018)Siam, Jiang, Lu, Petrich, Gamal, Elhoseiny, and
  Jagersand]{siam2018video}
Mennatullah Siam, Chen Jiang, Steven Lu, Laura Petrich, Mahmoud Gamal, Mohamed
  Elhoseiny, and Martin Jagersand.
\newblock Video segmentation using teacher-student adaptation in a human robot
  interaction ({HRI}) setting.
\newblock \emph{arXiv preprint arXiv:1810.07733}, 2018.

\bibitem[Simonyan and Zisserman(2014)]{2stream_cnn_simonyan_2014two}
Karen Simonyan and Andrew Zisserman.
\newblock Two-stream convolutional networks for action recognition in videos.
\newblock In \emph{NIPS}, 2014.

\bibitem[Song et~al.(2018)Song, Wang, Zhao, Shen, and Lam]{Song_2018_ECCV}
Hongmei Song, Wenguan Wang, Sanyuan Zhao, Jianbing Shen, and Kin-Man Lam.
\newblock Pyramid dilated deeper convlstm for video salient object detection.
\newblock In \emph{The European Conference on Computer Vision (ECCV)},
  September 2018.

\bibitem[Tokmakov and Alahari(2017)]{tokmakov2017learning}
Pavel Tokmakov and Karteek Alahari.
\newblock {Learning Video Object Segmentation with Visual Memory}.
\newblock In \emph{Proceedings of the IEEE International Conference on Computer
  Vision}, 2017.

\bibitem[Tokmakov et~al.(2017)Tokmakov, Alahari, and
  Schmid]{tokmakov2016learning}
Pavel Tokmakov, Karteek Alahari, and Cordelia Schmid.
\newblock Learning motion patterns in videos.
\newblock \emph{IEEE Conference on Computer Vision and Pattern Recognition},
  2017.

\bibitem[Tokmakov et~al.(2019)Tokmakov, Schmid, and
  Alahari]{tokmakov2019learning}
Pavel Tokmakov, Cordelia Schmid, and Karteek Alahari.
\newblock Learning to segment moving objects.
\newblock \emph{International Journal of Computer Vision}, 127\penalty0
  (3):\penalty0 282--301, 2019.

\bibitem[Tran et~al.(2015)Tran, Bourdev, Fergus, Torresani, and Paluri]{c3d}
Du~Tran, Lubomir Bourdev, Rob Fergus, Lorenzo Torresani, and Manohar Paluri.
\newblock Learning spatiotemporal features with 3d convolutional networks.
\newblock In \emph{Proceedings of the IEEE International Conference on Computer
  Vision}, 2015.

\bibitem[Tran et~al.(2018)Tran, Wang, Torresani, Ray, LeCun, and
  Paluri]{r2plus1d_cvpr18}
Du~Tran, Heng Wang, Lorenzo Torresani, Jamie Ray, Yann LeCun, and Manohar
  Paluri.
\newblock A closer look at spatiotemporal convolutions for action recognition.
\newblock In \emph{IEEE Conference on Computer Vision and Pattern Recognition},
  2018.

\bibitem[Voigtlaender and Leibe(2017)]{Voigtlaender2017Online}
Paul Voigtlaender and Bastian Leibe.
\newblock Online adaptation of convolutional neural networks for the 2017
  {DAVIS} challenge on video object segmentation.
\newblock In \emph{The 2017 DAVIS Challenge on Video Object Segmentation-CVPR
  Workshops}, 2017.

\bibitem[Xie et~al.(2018)Xie, Sun, Huang, Tu, and Murphy]{xie2018rethinking}
Saining Xie, Chen Sun, Jonathan Huang, Zhuowen Tu, and Kevin Murphy.
\newblock Rethinking spatiotemporal feature learning: Speed-accuracy trade-offs
  in video classification.
\newblock In \emph{Proceedings of the European Conference on Computer Vision
  (ECCV)}, pages 305--321, 2018.

\bibitem[Yan et~al.(2008)Yan, Khan, and Shah]{yan2008learning}
Pingkun Yan, Saad~M Khan, and Mubarak Shah.
\newblock Learning 4d action feature models for arbitrary view action
  recognition.
\newblock In \emph{2008 IEEE Conference on Computer Vision and Pattern
  Recognition}, pages 1--7. IEEE, 2008.

\bibitem[Yu and Koltun(2016)]{yu2015multi}
Fisher Yu and Vladlen Koltun.
\newblock Multi-scale context aggregation by dilated convolutions.
\newblock In \emph{ICLR}, 2016.

\bibitem[Yue-Hei~Ng et~al.(2015)Yue-Hei~Ng, Hausknecht, Vijayanarasimhan,
  Vinyals, Monga, and Toderici]{lstm_ng}
Joe Yue-Hei~Ng, Matthew Hausknecht, Sudheendra Vijayanarasimhan, Oriol Vinyals,
  Rajat Monga, and George Toderici.
\newblock Beyond short snippets: Deep networks for video classification.
\newblock In \emph{IEEE Conference on Computer Vision and Pattern Recognition},
  2015.

\bibitem[Zhao et~al.(2017)Zhao, Shi, Qi, Wang, and Jia]{zhao2017pyramid}
Hengshuang Zhao, Jianping Shi, Xiaojuan Qi, Xiaogang Wang, and Jiaya Jia.
\newblock Pyramid scene parsing network.
\newblock In \emph{Proceedings of the IEEE conference on computer vision and
  pattern recognition}, pages 2881--2890, 2017.

\end{thebibliography}
\end{document}